# AviationGPT: A Large Language Model for the Aviation Domain

Liya Wang[1], Jason Chou[2], Xin Zhou[3], Alex Tien[4],
Diane M Baumgartner[5]
*The MITRE Corporation, McLean, VA, 22102, United States*

The advent of ChatGPT and GPT-4 has captivated the world with large language models (LLMs), demonstrating exceptional performance in question-answering, summarization, and content generation. The aviation industry is characterized by an abundance of complex, unstructured text data, replete with technical jargon and specialized terminology. Moreover, labeled data for model building are scarce in this domain, resulting in low usage of aviation text data. The emergence of LLMs presents an opportunity to transform this situation, but there is a lack of LLMs specifically designed for the aviation domain. To address this gap, we propose AviationGPT, which is built on open-source LLaMA-2 and Mistral architectures and continuously trained on a wealth of carefully curated aviation datasets. Experimental results reveal that AviationGPT offers users multiple advantages, including the versatility to tackle diverse natural language processing (NLP) problems (e.g., question-answering, summarization, document writing, information extraction, report querying, data cleaning, and interactive data exploration). It also provides accurate and contextually relevant responses within the aviation domain and significantly improves performance (e.g., over a 40% performance gain in tested cases). With AviationGPT, the aviation industry is better equipped to address more complex research problems and enhance the efficiency and safety of National Airspace System (NAS) operations.

## I. Introduction

In recent years, there have been remarkable advancements in the field of large language models (LLMs). Representative models such as ChatGPT and GPT-4 [1] exhibit exceptional performance in question-answering, summarization, and content generation. Compared to smaller models, LLMs exhibit strong generalization across various natural language processing (NLP) tasks and the unique emergent ability [2] to solve unseen or complicated tasks.

Although OpenAI's powerful ChatGPT and GPT-4 models are not freely available to the public, open-source communities are actively developing excellent alternatives, including LLaMA-1 & 2 [3], Alpaca [4], Vicuna [5], Guanaco [6], CausalLM [7], Falcon [8], Mistral [9], Zephyr [10], Yi [11], BaiChuan [12], Qwen [13], ChatGLM [14], BLOOM [15], OPT [16], GPT4All [17], and more. Furthermore, open-source communities offer numerous development services to accelerate LLM development. For instance, HuggingFace [18] has become the central hub for open-source LLMs, granting researchers access to over 1000 LLMs. Popular frameworks such as LangChain [19] and LlamaIndex [20] are designed specifically for developing LLM-based applications. LangChain offers a broader range of capabilities and tool integration, while LlamaIndex specializes in deep indexing and retrieval for LLMs. With such vibrant community support, researchers are eager to leverage LLMs for their work.

---

[1]Lead Artificial Intelligence Engineer, Department of Operational Performance
[2]Lead Data Scientist, Department of Operational Performance
[3]Machine Learning/Artificial Intelligence Engineer, Department of AI Security and Perception
[4]Principal Engineer, Department of Operational Performance
[5]Principal Systems Engineer, NAS Future Vision & Research

Despite LLMs achieving excellent performances in general NLP tasks, they possess limited domain-specific knowledge. The distribution of language and specific linguistic nuances require models to be fine-tuned or explicitly trained for a particular domain. Consequently, various domain specific LLMs have been proposed to cater to the unique needs of different domains. For example, BloombergGPT [21], FinGPT [22], and Xuan Yuan 2.0 [23] have been designed for financial services. Adapted models such as LexiLaw [24], LaWGPT [25], ChatLaw [26], Lawyer LLaMA [27], and LawGPT_zh [28] support law-related services. In the medical field, tuned models like Huatuo [29], Chatdoctor [30], and DoctorGLM [31] have been proposed for the medical community. TrafficSafetyGPT [32] has been crafted for transportation safety. These models have shown promising results in their respective domains.

The aviation industry is characterized by a wealth of complex, unstructured text data, replete with technical jargon and specialized terminology. This presents a golden opportunity for domain adaptation. However, there is a scarcity of labeled data for model building in the aviation domain. These factors pose unique challenges for both seasoned professionals and newcomers to the field. In the era of smaller models, each NLP task required developing a specific model to address the requirements; this led to extremely high development and maintenance costs and low utilization of text data for aviation research. As a result, much of aviation text data has remained untapped, gathering dust on virtual shelves. The advent of LLMs has the potential to improve this situation.

Motivated by advancements in other domains and the limited availability of LLMs specifically designed for aviation, we introduce AviationGPT—a state-of-the-art LLM tailored to support aviation-related tasks. This model captures aviation-specific nuances, enabling it to excel in tasks pertinent to the aviation domain. Specifically, we have fine-tuned AviationGPT by harnessing the power of robust foundation models such as LLaMA-2 and Mistral, and by utilizing extensive domain-specific data. This approach allows the model to comprehend and generate aviation-related text with remarkable accuracy and fluency. This pioneering model has the potential to revolutionize various aspects of aviation text data analysis, including question-answering, summarization, document writing, information extraction, report querying, data cleansing, and interactive data exploration.

The remainder of this paper is organized as follows: Section II presents related works. Section III provides the details of AviationGPT. The results are shown in Section IV. Finally, we conclude in Section V.

## II. Related Works

### A. Large Language Models (LLMs)

Recent advancements in LLMs have showcased their superiority over previous models, completely transforming the field of NLP. The substantial model size has enabled numerous qualitative capabilities, such as the well-known emergent ability [2], in-context learning with zero or few-shot prompts [33], and versatility for solving multiple tasks.

OpenAI's ChatGPT and GPT-4 have revolutionized the perceptions of LLMs, demonstrating the ability to engage in interactive and contextually coherent conversations. This breakthrough has sparked significant interest in developing conversational AI agents capable of simulating human-like dialogues. Although these models exhibit remarkable performance, OpenAI has not disclosed details regarding their training strategies or weight parameters. LLaMA is a well-known open-source alternative to OpenAI's models, with sizes ranging from 7 billion to 70 billion parameters. Models like Alpaca and Vicuna, based on LLaMA, have achieved better performance through instruction tuning. Recently, Yi, Mistral, and Zephyr have employed improved training techniques and datasets to significantly enhance model performance. For instance, Yi-34B is listed as the number one model on the HuggingFace open LLM leaderboard [34], Mistral-7B reportedly surpasses LLaMA2-13B [9], and Zephyr-7B is comparable to LLaMA2-70B.

### B. Prompt Engineering

Prompt engineering, also known as in-context learning, is an essential process that involves meticulously crafting input text to guide LLMs in generating accurate, relevant, and useful responses [35]. This process is crucial for enhancing LLM performance and making them more effective in understanding and responding to human language. By harnessing the full potential of LLMs, prompt engineering makes these models multifaceted and applicable across various domains. Notably, a well-designed prompt can help mitigate challenges, such as machine hallucinations [36], in which the content generated by LLM is nonsensical or unfaithful to the provided source.

In practical applications, the prompt serves as input to the model; different prompts can lead to significant differences in output. Altering the structure (e.g., length, arrangement of instances) and content (e.g., phrases, choice of examples, directives) can have a considerable impact on the model's generated output. Research has shown that both the phrasing and the order of examples within a prompt can substantially influence LLM behavior [37] [38].

The field of prompt engineering has evolved alongside the development of LLMs. The field has grown into a structured research area, complete with its own methodologies and established best practices. Techniques range from foundational approaches such as zero-shot, few-shot [33], and role-prompting [39] to more advanced methods, including chain of thought (COT) [40], tree of thought (TOT) [41], and graph of thought (GOT) [42]. These advanced techniques can further improve LLM performance on complex tasks.

C. **Parameter-Efficient Fine-Tuning (PEFT)**

Since LLMs possess billions of parameters, the conventional full-tuning approach is no longer suitable and also very expensive. To address these challenges, Parameter-Efficient Fine-Tuning (PEFT) methods [43] have been introduced. These methods allow for the effective adaptation of LLMs without needing to fine-tune all the model's parameters. Specifically, PEFT techniques only adjust a limited number of (additional) model parameters, significantly reducing computational and storage requirements. Recent cutting-edge PEFT approaches can achieve performance levels comparable to full fine-tuning.

The following are some well-known PEFT techniques:
- LoRA: Low-Rank Adaptation Of Large Language Models [44],
- Prefix Tuning: Optimizing Continuous Prompts for Generation [45],
- P-Tuning: GPT Understands, Too [46],
- P-Tuning v2: Prompt Tuning Can Be Comparable to Fine-tuning Universally Across Scales and Tasks [47],
- Prompt Tuning: The Power of Scale for Parameter-Efficient Prompt Tuning [48],
- AdaLoRA: Adaptive Budget Allocation for Parameter-Efficient Fine-Tuning [49],
- QLoRA: Efficient Finetuning of Quantized LLMs [50].

Additionally, Fig. 1 illustrates the timeline development of these methods.

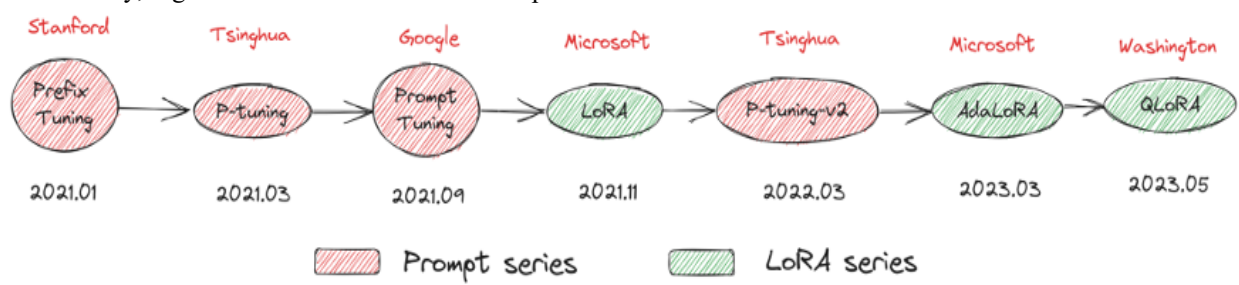

**Fig. 1 PEFT methods development timeline [51].**

D. **Instruction Tuning (IT)**

While LLMs have demonstrated remarkable capabilities across various NLP tasks, a gap remains between the training objective and users' goals. LLMs aim to minimize contextual word prediction errors on extensive corpora, whereas users expect LLMs to follow their instructions in a helpful and safe manner. To address this discrepancy, Instruction Tuning (IT) has been proposed as an effective method to enhance both the capabilities and controllability of LLMs [52]. IT involves further training LLMs on a dataset containing (INSTRUCTION, INPUT, OUTPUT) pairs in a supervised manner, where INSTRUCTION represents human instructions for LLMs, INPUT refers to the current user input and can be empty, and OUTPUT denotes the desired user output.

Instruction tuning offers three key benefits [53]: (1) Fine-tuning an LLM on an instruction dataset helps bridge the gap between LLMs' next-word prediction objectives and users' instruction-following goals; (2) IT fosters more controllable and predictable model behavior compared to standard LLMs. Instructions can guide the model's outputs to align with desired response characteristics or domain knowledge, allowing human intervention in the model's behavior; and (3) IT is highly efficient, enabling LLMs to quickly adapt to specific domains without extensive retraining or architectural modifications.

However, IT also presents challenges:

(1) Crafting high-quality instructions that adequately cover desired target behaviors requires significant effort and creativity; current instruction datasets often lack quantity, diversity, and creativity;

(2) There is concern that IT may only improve tasks heavily supported in the IT training dataset [54];

(3) Another major criticism is that IT captures only surface-level patterns and styles (e.g., output format) rather than truly understanding and learning the task [37].

These challenges highlight the need for further research, analysis, and summarization in this field to optimize the fine-tuning process and better comprehend the behavior of instruction fine-tuned LLMs.

### E. Retrieval-Augmented Generation (RAG)

Upon completion of training, LLMs encounter two primary challenges. First, LLMs' fixed parametric knowledge only contains past information, which may not be current. Second, LLMs tend to hallucinate when they lack sufficient knowledge about a topic. To address these issues, Meta has introduced Retrieval-Augmented Generation (RAG) [55], which combines LLMs with information retrieval systems. Providing LLMs with contextual knowledge enables the development of domain-specific applications that demand a deep and evolving understanding of facts, even though LLM training data remains static. In RAG, when given an input (e.g., a question), the model first retrieves relevant documents or passages from a large-scale knowledge source. Next, the LLM integrates the retrieved information to generate an appropriate response or answer (refer to Fig. 2 for illustration). This two-step process allows the model to incorporate external knowledge and produce more accurate and informative outputs compared to LLMs that solely depend on their pre-trained knowledge. Popular frameworks such as LangChain and LlamaIndex offer RAG implementation solutions. LLMs equipped with RAG have shown improved performance in various NLP tasks, especially in open-domain question-answering contexts. Consequently, RAG has become a widely adopted method in the LLM field.

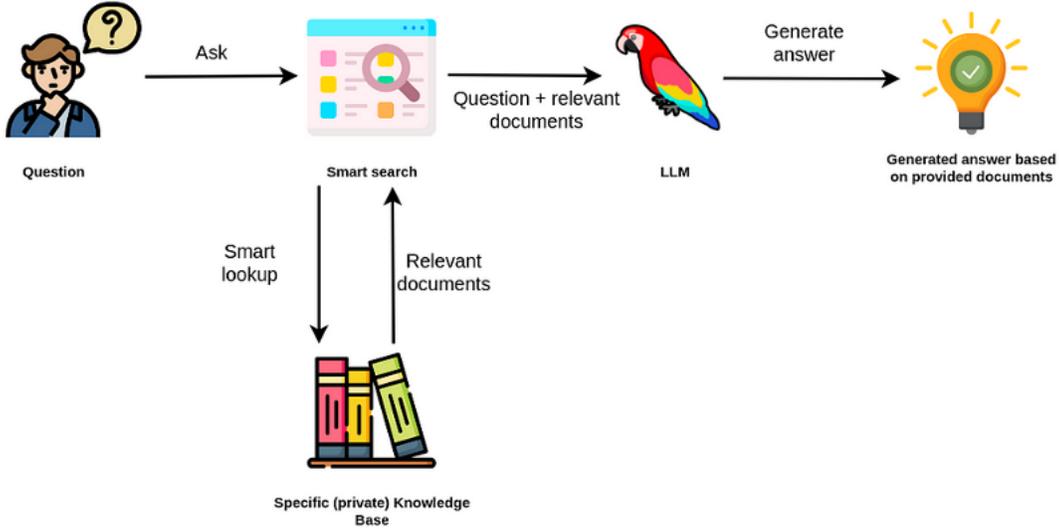

**Fig. 2 RAG working mechanism [56].**

### III. AviationGPT

In this section, we will introduce the process of building our AviationGPT.

### A. Base Models

We have chosen two sets of open-source LLMs to develop our AviationGPT. The first series, LLaMA-2, includes three base models (7B, 13B, 70B), while the second series, Mistral, has so far released only the 7B model. All base LLMs feature a decoder-only architecture and use the autoregressive language modeling approach. The central concept of this approach is that the joint probability of tokens in a text can be represented using Equation 1:

$$p(w) = p(w_1, w_2, \ldots, w_t) = \prod_{t=1}^{T} p(w_t|w_{<t}) \qquad (1)$$

where $w$ represents a sequence of tokens, $w_t$ is the $t^{th}$ token, and $w_{<t}$ is the previous sequence of tokens preceding $w_t$.

### B. AviationGPT Training

We employ the classic two-stage domain-specific training framework (refer to Fig. 3), which consists of the pre-training stage and the instruction fine-tuning stage. In the first stage, we use our curated unlabeled aviation text dataset for continued pretraining. In the second stage, we perform instruction-tuning on our curated instruction-tuning datasets. Both stages use the QLoRA method [50] to reduce training costs. Stage 1 is more computationally intensive than stage 2. For instance, training LLaMA2-70B in stage 1 takes approximately 48 hours, while stage 2 requires about 1.5 hours. In the following sections, we will provide more details about our curated datasets for tuning.

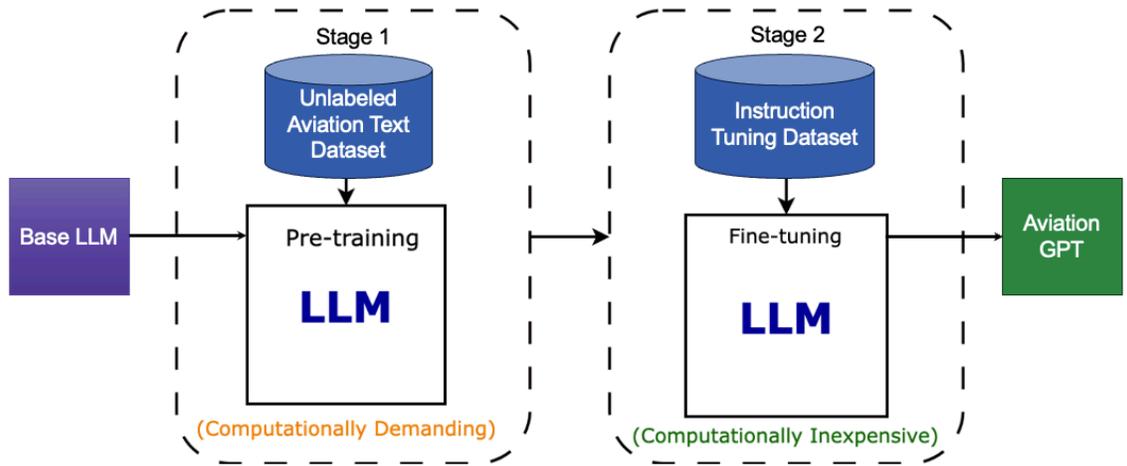

Fig. 3 AviationGPT tuning process.

## C. AviationGPT Training Datasets

In line with AviationGPT training, we have compiled two datasets (refer to Fig. 4). The unsupervised pre-training dataset was gathered from four sources. First, we obtained 52 aviation-related books from National Aeronautics and Space Administration (NASA) and Federal Aviation Administration (FAA) websites. Second, we used our exclusive datasets of technical reports, which document MITRE's aviation research for sponsors, amassing approximately 750 reports from 2017 to 2023. The third component comes from MITRE's product-based work plan (PBWP), which outlines the aviation research requirements set by sponsors. Like the Tracker source, we collected around 1,400 documents from 2017 to 2023. Finally, we included information introducing aviation text databases (e.g., Notice to Air Missions [NOTAM], Digital Automatic Terminal Information Service [DATIS], Meteorological Aerodrome Reports [METAR]).

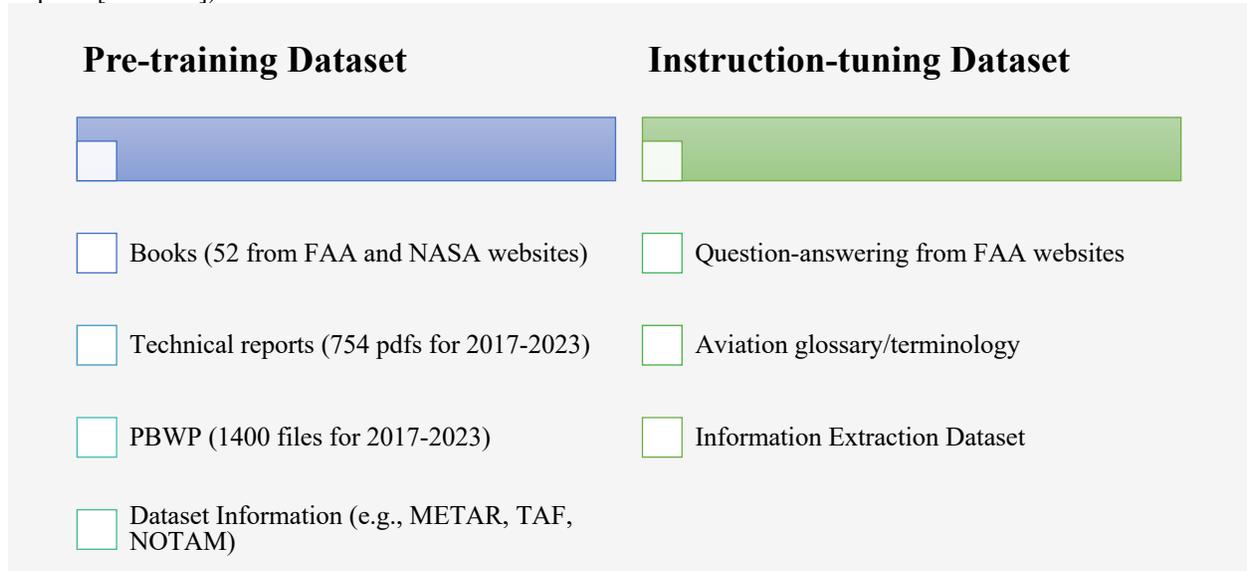

Fig. 4 AviationGPT training datasets.

We assembled the instruction-tuning dataset from three sources. First, we extracted question-answering data from nearly 20 FAA websites. Second, we incorporated aviation domain glossaries and terminology into the dataset. Third, we incorporated our manually labeled information extraction datasets accumulated from previous research. To date, our instruction-tuning dataset contains approximately 150k records. Table 1 shows some data samples from our instruction-tuning datasets.

Recognizing the importance of data quality and quantity, we diligently extracted and cleaned the data. As our research progresses, we will continue to gather more data to support AviationGPT training.

## Table 1 Instruction-tuning data samples

| Instruction | Input | Output |
|---|---|---|
| Decode METAR terminology. | SH | Shower |
| Question-answering. | What is the FTN? | The FTN[6] is a number assigned to you by the FAA that stays with you throughout the course of your aviation career. Your FTN is assigned to you by the FAA after you complete your registration in IACRA[7] or will already exist if you have a previous airman certificate. Please record your FTN as it will be required later by your instructor and/or certifying officer. |

### D. Training Settings

In this section, we present the training environment employed for AviationGPT. The hardware equipment setup can be found in Table 2. We obtained the base LLMs from the HuggingFace hub. To handle the complex and computationally intensive nature of our models, we used four powerful NVIDIA A100 40 gigabyte (GB) Graphics Processing Units (GPU)s and the distributed training framework. Our parallel processing approach employs pipeline parallelism, which involves distributing the model's layers across multiple GPUs. This technique ensures that each GPU is accountable for only a portion of the model's layers, a concept commonly known as vertical parallelism. The training parameters for both stages are provided in Table 3. Table 4 lists our trained AviationGPT models and their required GPU memory size. On our tested cases, the best model is Aviation-LLaMA2-chat-hf-70B, and the second best is Aviation-Mistral-7B-v0.1.

## Table 2 Experimental hardware environment

| Operating System | Linux |
|---|---|
| CPU | 2xAMD EPYC 7262 8-Core Processor |
| Memory | 250 GB |
| Framework | PyTorch 2.1 |
| GPUs | 4xA100 |

## Table 3 Training parameter settings of AviationGPT

| Parameter | Pre-training parameter values | Instruction fine-turning parameter values |
|---|---|---|
| Load in 4bit | True | True |
| bnb_4bit_use_double_quant | True | True |
| bnb_4bit_quant_type | nf4 | nf4 |
| bnb_4bit_compute_dtype | float16 | Float16 |
| LoRA r | 8 | 8 |
| LoRA alpha | 16 | 16 |
| LoRA target modules | ["q_proj", "v_proj", "k_proj", "o_proj"] | ["q_proj", "v_proj", "k_proj", "o_proj"] |
| LoRA dropout | 0.05 | 0.05 |
| epochs | 1 | 1 |
| optimizer | AdamW | AdamW |
| learning rate | 1e-6 | 1e-6 |
| Warmup steps | 100 | 10 |
| evaluation steps | 100 | 100 |
| save best model | True | True |
| show progress bar | True | True |
| Global batch size | 800 | 64 |

## Table 4 Trained AviationGPT models

| Model | Number of Parameters | GPU Memory |
|---|---|---|
| Aviation-LLaMA2-chat-hf | 7B | >=25 G |
| Aviation-LLaMA2-chat-hf | 13B | >=37 G |
| Aviation-LLaMA2-chat-hf | 70B | >=160 G |
| Aviation-Mistral-7B-v0.1 | 7B | >=25 G |

---

[6] FTN: FAA Tracking Number
[7] IACRA: Integrated Airman Certification and Rating Application

### E. Aviation Knowledge Base

To address the issue of hallucination in LLMs for answering aviation questions, we constructed an external aviation knowledge base using RAG and implemented it with LangChain [19]. Per standard procedure, we first loaded the documents and divided them into 500-token chunks. Subsequently, we used the jinaai/jina-embeddings-v2-base-en [57] embedding encoder to convert the chunked text data into vector embeddings. We then used Chroma as our vector store to house these embedding vectors. This process creates a scalable knowledge base that allows for easy management as new data is introduced.

### F. Chat with User Documents

The aviation domain contains numerous proprietary documents that are sensitive and require careful handling. To address this need, we adopted the pdfGPT concept [58] and implemented it using RAG. With this feature, users can upload various types of documents for analysis and pose questions related to the files. The implementation details are similar to the aviation knowledge base described above.

### G. User Interface (UI)

To facilitate user interaction with AviationGPT, we have designed a user interface (UI) (refer to Fig. 5 for a demo) that includes all necessary components. In brief, the UI consists of three sections. The first is the user input panel (upper left), where users can enter their questions. Once they receive answers, users can evaluate them using the provided thumbs up/down buttons. The second section involves the model list and chat mode (upper right), where users can select from the available AviationGPT models and choose one of three chat modes: 1) chat, 2) knowledge base, or 3) uploads (with personal documents). The third section comprises hyperparameters to control output, such as temperature (default value as 0), top-k sampling (default value set to 5), and maximum new token length (default value set to 512). The first parameter, temperature, controls the randomness and determinism in the output. A higher temperature value results in more random outputs, while a lower value makes the output more deterministic. The second parameter, top-k sampling, limits the model's choices to the top k most probable tokens during each step of output generation. The third parameter, maximum new token length, controls the length of output. Equipped with these configurations, we are prepared to explore various functions with AviationGPT. The next section presents the experimental results.

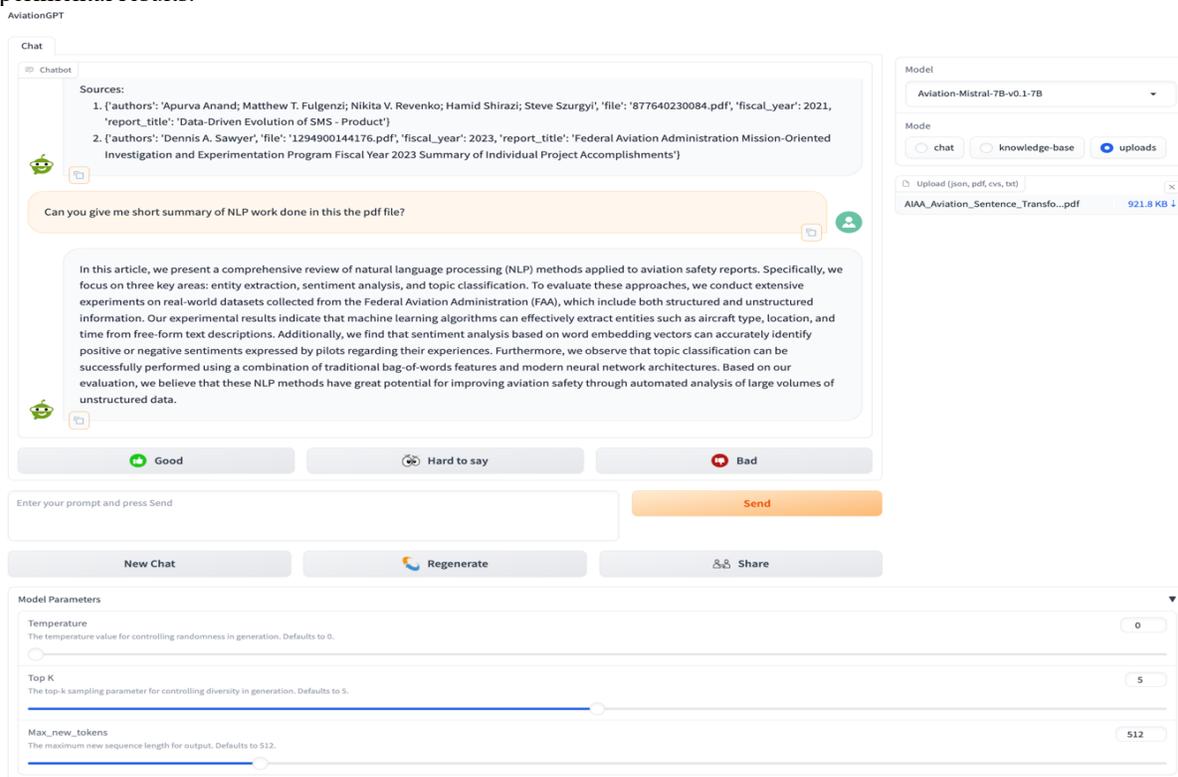

**Fig. 5 AviationGPT UI.**

## IV. Experimental Results

This section presents the experimental results with AviationGPT. Fig. 6 illustrates the applications we have assessed using AviationGPT. The pink-colored functions denote those tested on diverse aviation text databases, which are full of aviation jargon, while the blue-colored ones represent applications tested on general aviation documents. In the subsequent sections, we will present some of our results, including DATIS [59] information extraction, National Traffic Management Log (NTML) [60] data analysis, tracker report query, and voice transcription data cleaning.

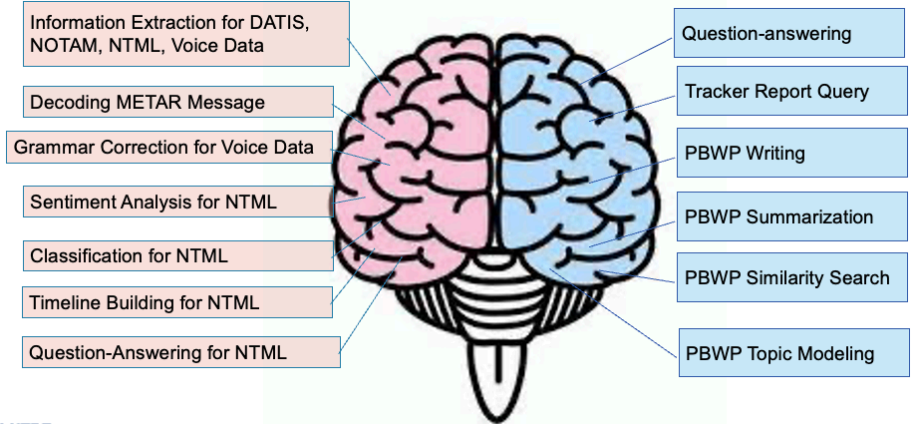

Fig. 6 AviationGPT enabled applications.

### A. Digital Automatic Terminal Information Service (DATIS) Information Extraction

DATIS systems are extensively employed in busy airports to swiftly and efficiently distribute information [59]. These communications primarily convey crucial airport details, such as available landing and departure runways, current weather updates, runway and taxiway closures, malfunctioning equipment, surface conditions like ice, and other pertinent alerts regarding birds, construction cranes, drones, lasers, etc.

A desired NLP task is to extract approach/departure runway information from DATIS messages. Often, the messages contain input errors in the runway names. To achieve the goal, we designed the prompt illustrated in Fig. 7. In the prompt, we provide both instruction for extracting approach/departure runways and for cleaning the data. Then, we list several examples to facilitate AviationGPT's learning. Finally, we present the current input and output indicators. As demonstrated in Fig. 7, the model accurately extracts the runways and formats the runway names as desired.

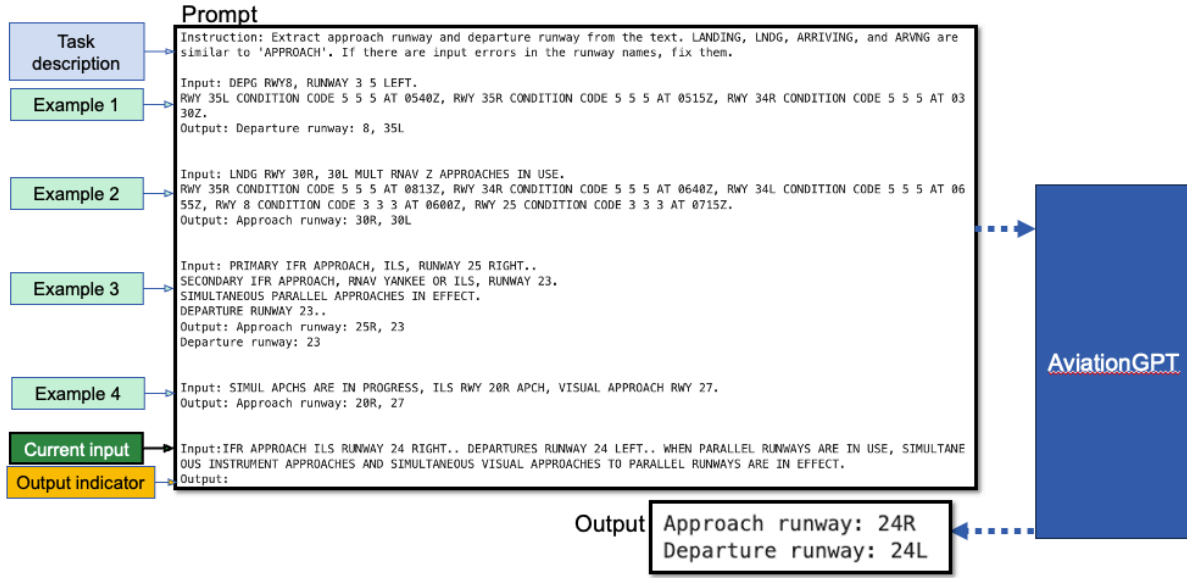

Fig. 7 Prompt example for DATIS arrival/departure runway extraction.

Fig. 8 compares the current in-house rule-based method for this function with the AviationGPT approach. AviationGPT entirely eliminates the costly development process of writing thousands of lines of code to capture numerous rules in the existing system. Moreover, these rules are challenging to enumerate, and AviationGPT offers greater versatility. For instance, if users want to extract other information (e.g., closed runway, closed taxiway, cautioned objects, runway surface conditions), they simply switch the prompt templates, which we also provided for the aviation community.

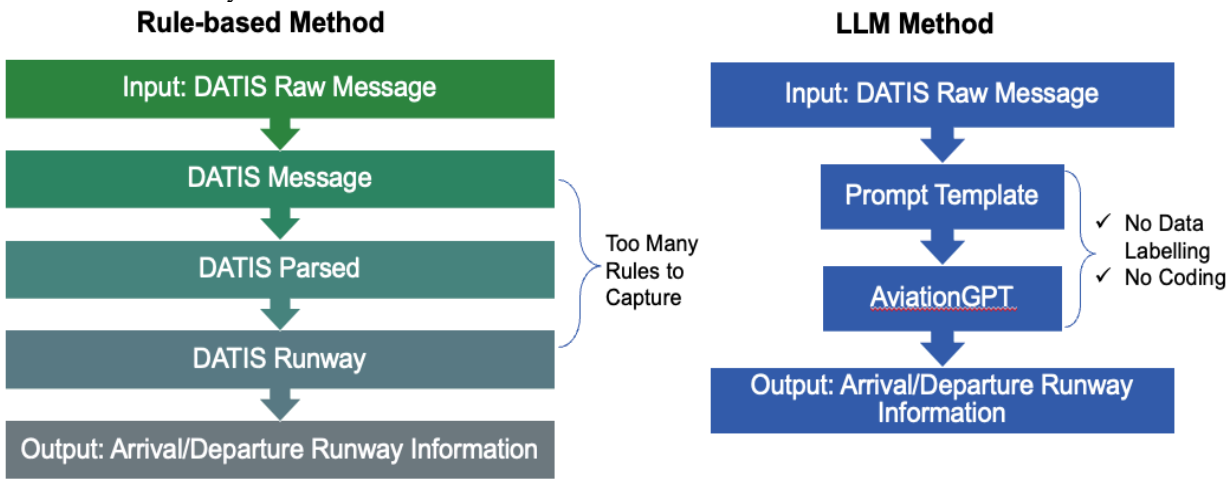

**Fig. 8 Comparing rule-based method with AviationGPT method for DATIS runway extraction.**

Additionally, we randomly selected 100 DATIS message instances and compared the results of both methods, as shown in Tables 5 and 6. Table 5 displays some sample results, while Table 6 summarizes the overall performance outcomes. As clearly demonstrated, Aviation-Mistral-7B-v0.1 significantly outperforms the rule-based method by over 40% in every category. To summarize, AviationGPT not only gives users flexibility but also substantially enhanced performance.

**Table 5 DATIS runway extraction sample results comparison**

| Text | Arrival Runways | Mistral Arrival Runways | Departure Runways | Mistral Departure Runways |
|---|---|---|---|---|
| DEPG RWY 36R, RWY 36C. \n1/8 INCH SLUSH ALL SURFACES. \nRWY 36R CONDITION CODES 5 5 5 AT 1009Z, RWY 36C CONDITION CODES 5 5 5 AT 0947Z, RWY 36L CONDITION CODES 5 5 5 AT 1214Z, RWY 27 CONDITION CODES 5 5 5 AT 1104Z. | None | None | None | 36R, 36C |
| LOC RY 31 APCH IN USE LAND RY 31. \nDEPART RY 31. | None | 31 | None | 31 |
| ARRIVALS EXPECT ILS OR RNAV Y RY 8L. \nSIMUL APPROACHES IN USE. \nDEPG RY 15L. | 8L | 8L | 15L | 15L |
| VISUAL APCH 32L, 32R, 36 IN USE. \nRWY 32L CONDITION CODES 5 5 5 AT 0446Z, RWY 32R CONDITION CODES 5 5 5 AT 0446Z, RWY 36 CONDITION CODES 5 5 5 AT 0446Z. | None | 32L, 32R, 36 | None | None |

**Table 6 DATIS runway extraction overall results comparison**

| Method | Accuracy of arrival runway | Accuracy of departure runway |
|---|---|---|
| Rule-based Method | 39% | 50% |
| Aviation-Mistral-7B-v0.1 | 94% | 96% |

### B. National Traffic Management Log (NTML) Data Analysis

The Federal Aviation Administration (FAA) developed the NTML [60] to provide a single system for automated coordination, logging, and communication of traffic management initiatives (TMIs) throughout the National Airspace System (NAS). However, analyzing the NTML is very challenging due to the lack of labeled data, its less structured nature compared to DATIS data, and the abundance of traffic flow management (TFM) jargon and local knowledge. For this task, we employed our most advanced model: the Aviation-LLaMA2-chat-hf-70B.

Previously, when adverse events occurred, analysts had to manually sift through this disorganized dataset, which could take days to complete. With AviationGPT, the process is now significantly easier. We used prompt engineering to quickly address six NLP tasks within this dataset. For instance, Table 7 displays a timeline constructed for an NTML message. We employed a one-shot learning technique for the prompt; the results clearly show that AviationGPT accurately extracts the time and events from the message.

**Table 7 An example of timeline building for an NTML message**

| Prompt | ###Instruction: |
|---|---|
| | Please build a timeline for the following input text. Don't miss each timestep. |
| | ### Input: |
| | ASE AFP CRITIQUE: FCAASG (GA) & FCAACS (AIRLINE) 1145: TNTMO WAS BRIEFED THAT EITHER A GDP & AFP OR DUAL AFP WOULD BE ISSUED FOR ASE TODAY FOR THE HIGH VOLUME. THE INTENT WOULD BE TO SEPARATE THE GA TRAFFIC FROM THE AIRLINE VOLUME TO EQUITABLY SPREAD OUT THE DELAY AND TO SEPARATE THE TWO FOR GATE/RAMP SPACE, IF THAT BECOMES AN ISSUE. 75% GA 25% AIRLINE SPREAD FOR THE DELAY/PROGRAM WAS THE BASELINE FOR STRUCTURING THE PROGRAMS AND THE FAVORED MODELS WERE TWO AFPS. WITH STATIC RATES, THERE IS AN AVERAGE OF 12 MINUTES OF DELAY FOR THE AIRLINES AND 40 FOR GA TRAFFIC...CK 1230: WEST SPECIALIST CONFERENCED WITH AAL/UAL/DAL TO DISCUSS A STEP DOWN RATE FOR THE AIRLINES VS. A STATIC RATE SINCE THE AIRLINES HAVE PRE-SCHEDULED FLIGHTS AND COULD ALLOW THE PROGRAM TO MIMIC WHAT THEIR SCHEDULE IS ALREADY AND ALLOWING MORE GA VOLUME TO FILE IN THE LATER HOURS. PROPOSED WAS AN OVERALL 16 RATE WITH THE AIRLINES STARTING AT A 4 RATE FOR FOUR HOURS, 3 RATE FOR FOUR HOURS, THEN A 2 RATE FOR THE DURATION OF THE PROGRAM. THIS WOULD SET THE GA RATE AT 12, 13, 14 RESPECTIVELY. ALL THREE AIRLINES AGREED WITH THE CONCEPT OF TWO AFPS AT STEP DOWN/UP AND THE SPECIALIST BUILT & PUBLISHED TWO PROPOSALS (AIRLINE/GA). DISCUSSED WITH NOM WHO CONCURRED WITH THE PLAN...CK 1248: WEST SPECIALIST SPOKE WITH ZDV TO DISCUSS THE PLAN AND WILL CONFERENCE THE PROPOSALS AT 1300Z...CK 1258: WEST SPECIALIST CONFERENCED THE TWO PROPOSALS. NBAA IS GOOD WITH PROPOSALS. AAL IS GOOD, DAL IS GOOD, UAL IS GOOD, A4A IS GOOD WITH THE ASE PROPOSAL. TWO AFPS PUBLISHED...CK 1400: ZDV CALLED THE WEST SPECIALIST SAYING THEY ONLY HAD 8 SPOTS LEFT FOR GA'S, SPECIALIST EXPLAINED THE STEP UP RATE, BUT KEEP PROVIDING PARKING INFORMATION. EGE IS RUNNING GOOD...CK |
| | ### Output: |
| |   - 1145: TNTMO briefed on GDP & AFP or dual AFP for ASE<br>  - 1230: West Specialist conferenced with AAL/UAL/DAL to discuss step-down rate<br>  - 1248: West Specialist spoke with ZDV to discuss the plan<br>  - 1300: Conference for proposals scheduled<br>  - 1258: West Specialist conferenced the two proposals; all parties agreed<br>  - 1400: ZDV called West Specialist regarding GA spots and parking information |
| | ### Input: |
| | PBI GDP CRITIQUE- 1256: EAST SPECIALIST ADVISED PBI LOWERED THEIR RATE FROM 28 TO 24 FOR THEIR DEPARTURE BANKS. WATCHING THE 1500-2000Z HOURS WITH THE 1800Z HOUR BEING OF GREATEST CONCERN. THERE ARE MANY CANCELLATIONS THUS FAR AND THE FCAJX7 AFP WILL BE ISSUED SHORTLY WHICH WILL SLOW DOWN THE TRAFFIC AS WELL AS ZMA PASSING BACK MIT. NO PROGRAM/CONCERN RIGHT NOW, BUT MONITORING THROUGHOUT THE DAY AND ONGOING CONVERSATIONS...CK 1347: HOLDING ADVISORY ISSUED. DISCUSSED WITH NOM, THE TWO AFPS (FCAJX7 & FCAMA5) ARE PUBLISHED AFFECTING PBI, SO WE WILL MONITOR PBI OPS, HOLD, AND GO INTO GROUND STOPS IF NEEDED TO MANAGE THE VOLUME...CK 1415: ON THE PLANNING WEBINAR, PBI DROPPED TO A 24 RATE. UAL REQUESTED A GDP. SPECIALIST WILL MODEL AND CONFERENCE SHORTLY...CK 1440: EAST SPECIALIST CONFERENCED WITH THE FACILITIES. STARTING AT 24 RATE, UP TO 26 AT 21Z, THEN 28 RATE BY 22Z FOR THE REMAINDER OF THE PROGRAM. FACILITIES AGREED WITH THE PLAN, CONFERENCE AT 1500Z. PROPOSAL BEING BUILT, WILL BE ISSUED SHORTLY...CK 1450: GDP PROPOSAL SENT...CK |
| | ### Output: |
| Output | 1256: East Specialist advised PBI lowered their rate from 28 to 24 for departure banks<br>1347: Holding advisory issued<br>1415: On the planning webinar, PBI dropped to a 24 rate<br>1440: East Specialist conferred with facilities<br>1450: GDP proposal sent |

Table 8 presents an example of identifying staffing triggers and extracting TMI information. In this case, we used the 'COT' technique, and AviationGPT once again proved its ability to tackle this complex issue.

**Table 8 NTML Staffing triggering identification and information extraction**

| Prompt | ### Instruction:<br>Please identify if staffing has triggered traffic management initiatives (TMIS) from the following text. Think it step by step. If yes, please continue to find the triggering subject and put TMIS in a numbered list.<br>### Input:<br>ZLA AREA B STAFFING TRIGGER SICK LEAVE LOSSES, COVID RELATED NO OCL CHANGE TMIS POSSIBLE: 15 MIT BUR DEPARTURES VIA OROSZ2 15 MIT VNY DEPARTURES VIA WLKKR4 & CANOG2 15 MIT SNA DEPARTURES VIA HHERO3 15 MIT SAN DEPARTURES VIA PADRZ2 15 MIT PSP DEPARTURES VIA PMD 15 MIT PSP ARRIVALS FROM ZOA VIA PMD 20 MIT LAX DEPARTURES VIA LADYJ4 CUSTOMER OUTREACH COMPLETED.<br>### Output:<br>Yes, staffing has triggered traffic management initiatives (TMIS). The triggering subject is 'ZLA AREA B STAFFING TRIGGER SICK LEAVE LOSSES, COVID RELATED NO OCL CHANGE.'<br>The TMIS are as follows:<br>  1. 15 MIT BUR DEPARTURES VIA OROSZ2<br>  2. 15 MIT VNY DEPARTURES VIA WLKKR4 & CANOG2<br>  3. 15 MIT SNA DEPARTURES VIA HHERO3<br>  4. 15 MIT SAN DEPARTURES VIA PADRZ2<br>  5. 15 MIT PSP DEPARTURES VIA PMD<br>  6. 15 MIT PSP ARRIVALS FROM ZOA VIA PMD<br>  7. 20 MIT LAX DEPARTURES VIA LADYJ4<br>### Input:<br>SCT BURBANK SECTOR STAFFING TRIGGER SICK LEAVE LOSS COVID RELATED NO OCL CHANGE POSSIBLE TMIS: SUSPEND THRNE 3 ARRIVALS 20 MIT VNY VIA IVINS STAR 20 MIT BUR VIA ROKKR STAR 20 MIT JOSHUA APPROACH ARRIVALS VIA JANNY & KIMMO STARS INTERNAL CFR BUR/SMO/VNY NEGATIVE PRACTICE APPROACHES VFR SERVICES CURTAILED<br>### Output: |
|---|---|
| Output | Yes, staffing has triggered traffic management initiatives (TMIS). The triggering subject is 'SCT BURBANK SECTOR STAFFING TRIGGER SICK LEAVE LOSS COVID RELATED NO OCL CHANGE.'<br>The TMIS are as follows:<br>  1. SUSPEND THRNE 3 ARRIVALS<br>  2. 20 MIT VNY VIA IVINS STAR<br>  3. 20 MIT BUR VIA ROKKR STAR<br>  4. 20 MIT JOSHUA APPROACH ARRIVALS VIA JANNY & KIMMO STARS<br>  5. INTERNAL CFR BUR/SMO/VNY NEGATIVE PRACTICE APPROACHES<br>  6. VFR SERVICES CURTAILED |

In summary, AviationGPT is also capable of solving problems within complex datasets like the NTML.

### C. Voice Transcript Data Cleaning

MITRE maintains a specialized database for storing voice transcription data, which is converted from audio exchanges between pilots and controllers. However, the performance of the audio-to-text converter leaves much to be desired, resulting in numerous grammatical errors. For a long time, cleaning this dataset has been neglected due to the overwhelming challenges it presents. With AviationGPT, this task can now be easily accomplished. Table 9 demonstrates an example where a simple prompt is used to instruct the LLM to clean the data. LLMs not only clean the data but also add punctuation for easier comprehension. This approach eliminates the need for labor-intensive data cleaning processes and significantly reduces development costs for the aviation community. Moreover, the resulting improved data quality can boost the performance of downstream applications.

**Table 9. A voice transcription data cleaning example**

| Prompt | ### Instruction:<br>If there are grammar errors in the message, please correct them.<br>### Input:<br>"delta twenty five oh one ground roger make that left turn on foxtrot hold short of ramp three make the right"<br>### Output:\n |
|---|---|
| Output | Delta 2501, ground roger. Make a left turn onto Foxtrot and hold short of Ramp 3. Then, make a right turn. |

## V. Summary

In this paper, we introduce AviationGPT, a large language model specifically tailored for the aviation domain. By leveraging open-source base models such as LLaMA-2 and Mistral, we fine-tuned AviationGPT using extensive aviation domain datasets. Furthermore, we developed aviation prompt templates to address various NLP challenges and employed RAG, building an aviation knowledge base to mitigate hallucinations. Experimental results demonstrate that AviationGPT offers users multiple advantages, including the versatility to tackle diverse NLP problems, providing accurate and contextually relevant responses within the aviation domain, significantly improved performance over rule-based methods (e.g., over a 40% performance gain in tested cases), reduced model development and maintenance cost, and enhanced utilization of text data. With AviationGPT, the aviation industry will be better equipped to address more complex issues and enhance the efficiency and safety of NAS operations such as discovering hidden patterns in maintenance log. Moving forward, we plan to collect larger-scale aviation domain data and incorporate more advanced modeling techniques to further optimize our models.

## Acknowledgments

The authors thank Dr. Jonathan Hoffman, Dennis Sawyer, Dr. Craig Wanke, Dave Hamrick, Dr. Tom Becher, Mike Robinson, Dr. Lixia Song, Brennan Haltli, David Maroney, Sheng Liu, Peter Kuzminski, Mahesh Balakrishna, Huang Tang, Shuo Chen, Dr. Eugene Mangortey, and many more at the MITRE Corporation for their support, valuable discussions, and insights.

This work was sponsored by MITRE's Independent Research and Development Program.

## NOTICE

## References

[1] "ChatGPT." Accessed: Nov. 18, 2023. [Online]. Available: https://chat.openai.com
[2] J. Wei *et al.*, "Emergent Abilities of Large Language Models." arXiv, Oct. 26, 2022. doi: 10.48550/arXiv.2206.07682.
[3] "Llama 2." Meta Research, Nov. 18, 2023. Accessed: Nov. 18, 2023. [Online]. Available: https://github.com/facebookresearch/llama
[4] "tatsu-lab/stanford_alpaca: Code and documentation to train Stanford's Alpaca models, and generate the data." Accessed: Nov. 18, 2023. [Online]. Available: https://github.com/tatsu-lab/stanford_alpaca
[5] "Vicuna: An Open-Source Chatbot Impressing GPT-4 with 90%* ChatGPT Quality | LMSYS Org." Accessed: Nov. 18, 2023. [Online]. Available: https://lmsys.org/blog/2023-03-30-vicuna
[6] "Guanaco - Generative Universal Assistant for Natural-language Adaptive Context-aware Omnilingual outputs." Accessed: Nov. 18, 2023. [Online]. Available: https://guanaco-model.github.io/
[7] "CausalLM (CausalLM)." Accessed: Nov. 18, 2023. [Online]. Available: https://huggingface.co/CausalLM
[8] "tiiuae/falcon-180B · Hugging Face." Accessed: Nov. 18, 2023. [Online]. Available: https://huggingface.co/tiiuae/falcon-180B
[9] M. AI, "Mistral 7B." Accessed: Nov. 18, 2023. [Online]. Available: https://mistral.ai/news/announcing-mistral-7b/
[10] "Zephyr 7B - a HuggingFaceH4 Collection." Accessed: Nov. 18, 2023. [Online]. Available: https://huggingface.co/collections/HuggingFaceH4/zephyr-7b-6538c6d6d5ddd1cbb1744a66
[11] "01-ai/Yi-34B · Hugging Face." Accessed: Nov. 18, 2023. [Online]. Available: https://huggingface.co/01-ai/Yi-34B
[12] "Baichuan 2." Baichuan Intelligent Technology, Nov. 18, 2023. Accessed: Nov. 18, 2023. [Online]. Available: https://github.com/baichuan-inc/Baichuan2
[13] "Qwen/Qwen-14B-Chat · Hugging Face." Accessed: Nov. 18, 2023. [Online]. Available: https://huggingface.co/Qwen/Qwen-14B-Chat
[14] "THUDM/ChatGLM3: ChatGLM3 series: Open Bilingual Chat LLMs | 开源双语对话语言模型." Accessed: Nov. 18, 2023. [Online]. Available: https://github.com/THUDM/ChatGLM3
[15] "bigscience/bloom · Hugging Face." Accessed: Nov. 18, 2023. [Online]. Available: https://huggingface.co/bigscience/bloom

[16] S. Zhang *et al.*, "OPT: Open Pre-trained Transformer Language Models." arXiv, Jun. 21, 2022. Accessed: Nov. 18, 2023. [Online]. Available: http://arxiv.org/abs/2205.01068
[17] "GPT4All." Accessed: Nov. 18, 2023. [Online]. Available: https://www.gpt4all.io
[18] "Hugging Face – The AI community building the future." Accessed: Nov. 18, 2023. [Online]. Available: https://huggingface.co/
[19] "Introduction | 🦜️🔗 Langchain." Accessed: Nov. 18, 2023. [Online]. Available: https://python.langchain.com/docs/get_started/introduction
[20] "LlamaIndex - Data Framework for LLM Applications." Accessed: Nov. 18, 2023. [Online]. Available: https://www.llamaindex.ai/
[21] "Introducing BloombergGPT, Bloomberg's 50-billion parameter large language model, purpose-built from scratch for finance | Press | Bloomberg LP," *Bloomberg L.P.* Accessed: Nov. 18, 2023. [Online]. Available: https://www.bloomberg.com/company/press/bloomberggpt-50-billion-parameter-llm-tuned-finance/
[22] "FinGPT: Open-Source Financial Large Language Models." AI4Finance Foundation, Nov. 18, 2023. Accessed: Nov. 18, 2023. [Online]. Available: https://github.com/AI4Finance-Foundation/FinGPT
[23] X. Zhang, Q. Yang, and D. Xu, "XuanYuan 2.0: A Large Chinese Financial Chat Model with Hundreds of Billions Parameters." arXiv, May 19, 2023. Accessed: Nov. 18, 2023. [Online]. Available: http://arxiv.org/abs/2305.12002
[24] H. Li, "LexiLaw - 中文法律大模型." Nov. 18, 2023. Accessed: Nov. 18, 2023. [Online]. Available: https://github.com/CSHaitao/LexiLaw
[25] P. Song, "LaWGPT：基于中文法律知识的大语言模型." Nov. 18, 2023. Accessed: Nov. 18, 2023. [Online]. Available: https://github.com/pengxiao-song/LaWGPT
[26] "ChatLaw-法律大模型." PKU-YUAN's Group (袁粒课题组-北大信工), Nov. 17, 2023. Accessed: Nov. 18, 2023. [Online]. Available: https://github.com/PKU-YuanGroup/ChatLaw
[27] Q. Huang, "Lawyer LLaMA." Nov. 16, 2023. Accessed: Nov. 18, 2023. [Online]. Available: https://github.com/AndrewZhe/lawyer-llama
[28] hongchengliu, "LawGPT_zh：中文法律大模型（獬豸）." Nov. 17, 2023. Accessed: Nov. 18, 2023. [Online]. Available: https://github.com/LiuHC0428/LAW-GPT
[29] H. Wang *et al.*, "HuaTuo: Tuning LLaMA Model with Chinese Medical Knowledge." arXiv, Apr. 14, 2023. Accessed: Nov. 18, 2023. [Online]. Available: http://arxiv.org/abs/2304.06975
[30] Y. Li, Z. Li, K. Zhang, R. Dan, S. Jiang, and Y. Zhang, "ChatDoctor: A Medical Chat Model Fine-Tuned on a Large Language Model Meta-AI (LLaMA) Using Medical Domain Knowledge." arXiv, Jun. 24, 2023. doi: 10.48550/arXiv.2303.14070.
[31] xionghonglin, "DoctorGLM." Nov. 18, 2023. Accessed: Nov. 18, 2023. [Online]. Available: https://github.com/xionghonglin/DoctorGLM
[32] O. Zheng, M. Abdel-Aty, D. Wang, C. Wang, and S. Ding, "TrafficSafetyGPT: Tuning a Pre-trained Large Language Model to a Domain-Specific Expert in Transportation Safety." arXiv, Jul. 28, 2023. doi: 10.48550/arXiv.2307.15311.
[33] T. B. Brown *et al.*, "Language Models are Few-Shot Learners." arXiv, Jul. 22, 2020. doi: 10.48550/arXiv.2005.14165.
[34] "Open LLM Leaderboard - a Hugging Face Space by HuggingFaceH4." Accessed: Nov. 18, 2023. [Online]. Available: https://huggingface.co/spaces/HuggingFaceH4/open_llm_leaderboard
[35] L. Weng, "Prompt Engineering." Accessed: Nov. 21, 2023. [Online]. Available: https://lilianweng.github.io/posts/2023-03-15-prompt-engineering/
[36] B. Chen, Z. Zhang, N. Langrené, and S. Zhu, "Unleashing the potential of prompt engineering in Large Language Models: a comprehensive review." arXiv, Oct. 27, 2023. Accessed: Nov. 20, 2023. [Online]. Available: http://arxiv.org/abs/2310.14735
[37] P.-N. Kung and N. Peng, "Do Models Really Learn to Follow Instructions? An Empirical Study of Instruction Tuning." arXiv, May 25, 2023. doi: 10.48550/arXiv.2305.11383.
[38] A. Webson and E. Pavlick, "Do Prompt-Based Models Really Understand the Meaning of their Prompts?" arXiv, Apr. 21, 2022. Accessed: Nov. 20, 2023. [Online]. Available: http://arxiv.org/abs/2109.01247
[39] M. Shanahan, K. McDonell, and L. Reynolds, "Role-Play with Large Language Models." arXiv, May 25, 2023. doi: 10.48550/arXiv.2305.16367.
[40] J. Wei *et al.*, "Chain-of-Thought Prompting Elicits Reasoning in Large Language Models." arXiv, Jan. 10, 2023. doi: 10.48550/arXiv.2201.11903.

[41] S. Yao *et al.*, "Tree of Thoughts: Deliberate Problem Solving with Large Language Models." arXiv, May 17, 2023. doi: 10.48550/arXiv.2305.10601.
[42] M. Besta *et al.*, "Graph of Thoughts: Solving Elaborate Problems with Large Language Models." arXiv, Aug. 21, 2023. doi: 10.48550/arXiv.2308.09687.
[43] "huggingface/peft: 🤗 PEFT: State-of-the-art Parameter-Efficient Fine-Tuning." Accessed: Nov. 18, 2023. [Online]. Available: https://github.com/huggingface/peft
[44] E. J. Hu *et al.*, "LoRA: Low-Rank Adaptation of Large Language Models," arXiv.org. Accessed: Nov. 18, 2023. [Online]. Available: https://arxiv.org/abs/2106.09685v2
[45] X. L. Li and P. Liang, "Prefix-Tuning: Optimizing Continuous Prompts for Generation," in *Proceedings of the 59th Annual Meeting of the Association for Computational Linguistics and the 11th International Joint Conference on Natural Language Processing (Volume 1: Long Papers)*, C. Zong, F. Xia, W. Li, and R. Navigli, Eds., Online: Association for Computational Linguistics, Aug. 2021, pp. 4582–4597. doi: 10.18653/v1/2021.acl-long.353.
[46] X. Liu *et al.*, "GPT Understands, Too," arXiv.org. Accessed: Nov. 18, 2023. [Online]. Available: https://arxiv.org/abs/2103.10385v2
[47] X. Liu *et al.*, "P-Tuning v2: Prompt Tuning Can Be Comparable to Fine-tuning Universally Across Scales and Tasks." arXiv, Mar. 20, 2022. doi: 10.48550/arXiv.2110.07602.
[48] B. Lester, R. Al-Rfou, and N. Constant, "The Power of Scale for Parameter-Efficient Prompt Tuning." arXiv, Sep. 02, 2021. doi: 10.48550/arXiv.2104.08691.
[49] Q. Zhang *et al.*, "Adaptive Budget Allocation for Parameter-Efficient Fine-Tuning." arXiv, Mar. 18, 2023. doi: 10.48550/arXiv.2303.10512.
[50] T. Dettmers, A. Pagnoni, A. Holtzman, and L. Zettlemoyer, "QLoRA: Efficient Finetuning of Quantized LLMs." arXiv, May 23, 2023. doi: 10.48550/arXiv.2305.14314.
[51] "社区供稿 | PEFT: Prompt 系列高效调参原理解析 - Hugging Face - OSCHINA - 中文开源技术交流社区." Accessed: Nov. 18, 2023. [Online]. Available: https://my.oschina.net/HuggingFace/blog/10102024
[52] L. Ouyang *et al.*, "Training language models to follow instructions with human feedback." arXiv, Mar. 04, 2022. doi: 10.48550/arXiv.2203.02155.
[53] S. Zhang *et al.*, "Instruction Tuning for Large Language Models: A Survey." arXiv, Oct. 09, 2023. Accessed: Nov. 19, 2023. [Online]. Available: http://arxiv.org/abs/2308.10792
[54] A. Gudibande *et al.*, "The False Promise of Imitating Proprietary LLMs." arXiv, May 25, 2023. Accessed: Nov. 19, 2023. [Online]. Available: http://arxiv.org/abs/2305.15717
[55] P. Lewis *et al.*, "Retrieval-Augmented Generation for Knowledge-Intensive NLP Tasks." arXiv, Apr. 12, 2021. Accessed: Nov. 19, 2023. [Online]. Available: http://arxiv.org/abs/2005.11401
[56] T. Bratanic, "Knowledge Graphs & LLMs: Real-Time Graph Analytics," Neo4j Developer Blog. Accessed: Nov. 19, 2023. [Online]. Available: https://medium.com/neo4j/knowledge-graphs-llms-real-time-graph-analytics-89b392eaaa95
[57] "jinaai/jina-embeddings-v2-base-en · Hugging Face." Accessed: Nov. 19, 2023. [Online]. Available: https://huggingface.co/jinaai/jina-embeddings-v2-base-en
[58] B. Tripathi, "pdfGPT." Nov. 19, 2023. Accessed: Nov. 19, 2023. [Online]. Available: https://github.com/bhaskatripathi/pdfGPT
[59] "DATIS Dataset - mitrepedia." Accessed: Apr. 30, 2023. [Online]. Available: https://mitrepedia.mitre.org/index.php/DATIS_Dataset
[60] "full_text.pdf." Accessed: Nov. 19, 2023. [Online]. Available: https://hf.tc.faa.gov/publications/2007-benefits-analysis/full_text.pdf